# Using Spark Machine Learning Models to Perform Predictive Analysis on Flight Ticket Pricing Data

Philip Wong, Phue Thant, Pratiksha Yadav, Ruta Antaliya, Jongwook Woo
Department of Information Systems, California State University Los Angeles
{pwong9, pthant, pyadav, rantali, jwoo5}@calstatela.edu

**Abstract:**
This paper discusses predictive performance and processes undertaken on flight pricing data utilizing r2(r-square) and RMSE that leverages a large dataset, originally from Expedia.com, consisting of approximately 20 million records or 4.68 gigabytes. The project aims to determine the best models usable in the real world to predict airline ticket fares for non-stop flights across the US. Therefore, good generalization capability and optimized processing times are important measures for the model.

We will discover key business insights utilizing feature importance and discuss the process and tools used for our analysis. Four regression machine learning algorithms were utilized: Random Forest, Gradient Boost Tree, Decision Tree, and Factorization Machines utilizing Cross Validator and Training Validator functions for assessing performance and generalization capability.

## 1. Introduction & Business Case

The business case for predicting pricing has two stakeholders' airlines operators and Customers. Airlines may optimize their pricing strategy by predicting pricing trends over time, setting appropriate prices for specific routes, and allowing the airline to compare future prices against their competitors, creating a competitive advantage. In the example of competitive advantage and pricing optimization, a low-cost carrier such as Southwest Airlines may want to set prices for future flights that are lower cost than their competitors but not too low to which the carrier will be leaving revenue on the table it could have generated. Likewise, travelers may find the analysis can provide unique insights into what specific factors (or features) can impact airline pricing. This would allow the savvy traveler to gain an advantage over others. Also, the ability to forecast ticket fares enables the consumer to better plan budgets and find opportunities to lower the overall cost of airfare.

## 2. Data Set Used

The dataset used in this paper is taken from an open-source platform which is Kaggle.com [5]. The dataset is of csv format, and contains information about Expedia's purchasable tickets between the months of April 2022 and October 2022.

|  | **Original** | **Cleaning - Full Dataset** | **Sample** |
|---|---|---|---|
| **Fields** | 27 | 15 | |
| **Records** | ~200m | ~20m | ~100k |
| **Size** | 31.09GB | 4.86GB | 23.4MB |

Fig 1 Dataset Specification

## 2.1 DATASET SUMMARY

The different columns present in the dataset are listed below. Note the label column used is the *baseFare*. There are a total of 14 features and 1 Label (*baseFare*). That dataset was filtered only to include flight legs for non-stop flights (20m records), in which a randomized sample of 10% was used for parameter tuning.

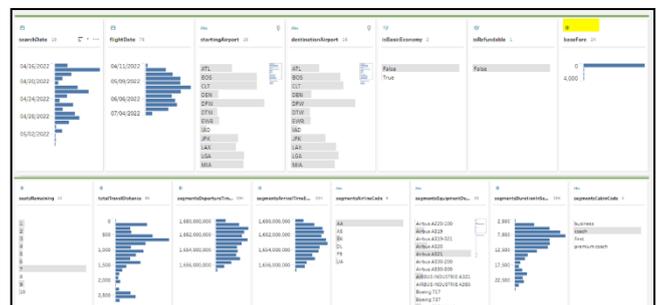

Fig 2 Features and Label Overview – TableauPrep

## 3. Technical Specifications

Table 1 H/W Specification

| Cluster Version | Hadoop 3.2.1-amzn-3.1 |
|---|---|
| No of CPUs | 8 CPUs |
| Pyspark Version | 3.0 |
| Number of Nodes | 5 |
| CPU speed | 2.20 GHz |
| Total Storage | 481 GB |

## 4. Related Work

The paper, "Airline Fare Prediction Using Machine Learning Algorithms", investigates the variables influencing airfare and aims to forecast travel costs [1]. It analyzes flight schedules, destinations, length, holidays, and vacations as factors affecting ticket prices. The study employs seven machine learning models, including linear quantile mixed regression, Learn++, Bayesian estimation with Kalman filter, and ARMA mixed with random forest algorithms. It also utilizes regression machine learning models like Extreme Learning Machine (ELM), Multilayer Perceptron (MLP), Generalized Regression Neural Network, Random Forest Regression Tree, Regression Tree, Linear Regression (LR), and Regression SVM (Polynomial) for cost prediction. Our research shares the same objective of comparing machine learning models for predicting airline ticket costs; however, it distinguishes itself by incorporating Factorization Machines (Fm) and Gradient Boosted Tree (GBT) techniques, which are not used in the mentioned paper.

The paper, "Flight Fare Prediction System Using Machine Learning"; explores the use of machine learning algorithms for predicting airline ticket prices [2]. The study compares various supervised learning algorithms, including Classification Tree (CART), Logistic Regression (LR), Naive Bayes, SoftMax Regression, and Support Vector Machines (SVMs), to classify ticket prices into different bins relative to the average price. The dataset encompasses various attributes, including the source and destination of flights, departure dates, departure times, number of stops, arrival times, and corresponding costs. This article aims to achieve the same objective as our research work which is to compare different machine-learning models for predicting airline ticket costs. In this article, SVM classification is used to categorize costs as "greater" or "lower" than the average, which is not used in our article.

The paper, "Regression – Flight Price Prediction," conducts an exploratory analysis of the data and performs similar regression testing using DT, SVR, KNN, and LR in addition to ensemble models [3]. There are 13,000 records in the dataset and 11 fields. Our research shares the same objective of comparing machine learning models for predicting airline ticket costs, and the paper also identified GBT as the best-performing algorithm; however, the difference includes critical differences. Firstly, the dataset difference in size and richness is substantially smaller and less complex compared to our dataset of 20 million records and 15 fields analyzed; our sample size alone of nearly 100k records is already much larger. We also leveraged Hadoop BigData systems to process our larger dataset while the author did not specify any technical information regarding the tools or systems used in their analysis. In terms of method, the author of this paper did not describe or use feature importance or engineering to determine fields or draw business insights similar to our analysis.

## 5. FEATURE IMPORTANCE AND ENGINEERING

| feature | importance |
| --- | --- |
| startingAirportIdx | 0.117477 |
| segmentsDurationInSeconds | 0.117477 |
| destinationAirportIdx | 0.117353 |
| seatsRemaining | 0.117353 |
| segmentsArrivalTimeEpochSeconds | 0.101249 |
| segmentsCabinCodeIdx | 0.101249 |
| segmentsEquipmentDescriptionIdx | 0.099129 |
| segmentsAirlineCodeIdx | 0.079707 |
| totalTravelDistance | 0.079707 |
| SearchMonth | 0.062302 |
| segmentsDepartureTimeEpochSeconds | 0.050651 |
| isBasicEconomy | 0.050651 |
| flightMonth | 0.037088 |
| SearchDay | 0.010182 |
| flightDay | 0.004921 |
| SearchYear | 0.000000 |
| isRefundable | 0.000000 |
| flightYear | 0.000000 |

Fig 3 Feature Importance of Gradient Boost Tree

Using feature importance generated with the GBT algorithm was used for our analysis of the features as it was one of the best-performing machine learning algorithms for the model.

### 5.1 FEATURE ENGINEERING FINDINGS
The features *StartingAiport* and *FlightDuration* both had the most impact on the model, followed by *DestinationAirport* and *SeatsRemaining*. This appears to be expected as these parameters will have a more significant impact on pricing. A key finding was the *AirlineEquipment* field, which represents the aircraft used for the specific flight segment in the context of the data. This field had a small but consistent impact on the R2 values across the machine-learning algorithms. This could mean that travelers looking to find a lower-cost alternative may be able to identify specific planes and imply that newer planes have low operating costs, leading to lower flight ticket prices.

Concatenating fields that may have more meaning together than separate such as depart and arrival airport (which creates a unique route), had little to no impact on R2. Another finding was despite the large dataset, the timeframe of the data was between 3 and 4 months which may have limited the predictive pricing performance of our models.

## 6. Workflow Architecture

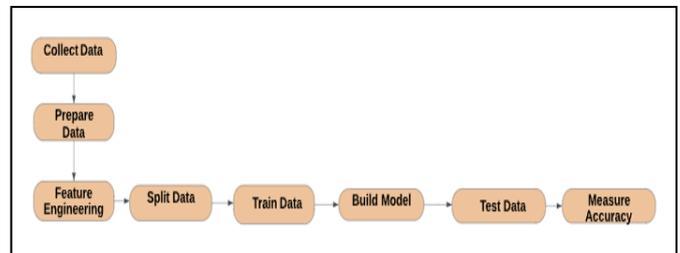

Fig 4 Workflow Architecture Diagram

Figure 4 shows the workflow to build the Machine Learning models. The first step was to find the right source of data. In this dataset of flights that we have chosen, we have defined objectives in terms of measurable goals that we wanted to achieve by the end of the project. We downloaded the dataset from the Kaggle website. Once the data is collected, it needs to be processed and prepared for the use of the model. This involved tasks like cleaning the data, scaling, and normalizing the features, and transforming the data into a format suitable for the model.

We also performed feature engineering to see which features have the highest feature importance and tried to see which features are appropriate for the model. Before training the model, it is important to split the data into a training set and a testing set. We have split 70% data into training data and 30% into testing data. The training set is to train the model, while the testing set is used to evaluate the performance of the model and ensure that it generalizes well to new, unseen data. Now that the data is been prepared and split, the next step is to train the machine-learning model using an appropriate algorithm.

It also involves finding the best parameters or hyperparameters to optimize the model's performance. Then we build a model with the trained dataset. Once the model has been trained, it needs to be tested and validated using the testing set. We implement models using regression algorithms so that the accuracy of the models can be measured through Root Mean Square Error (RMSE) and Coefficient of Determination (R2).

## 7. Machine Learning Algorithms
Regression is a supervised machine learning (ML) technique that is used to predict continuous values. In part one of our project involving flight price prediction, we make use of Regression models, as price, the target variable is a continuous numeric variable. We used four Regression algorithms - Random Forest Regression, Gradient Boost Tree Regression, Factorization Machine Regression, and Decision Forest Regression. We have implemented these models on the Sample Dataset to predict the flight price. These four algorithms were implemented in PySpark.

### 7.1 Random Forest Regression
The parameters that we have used in paramGrid are used to fine-tune the Random Forest model through hyperparameter tuning. For example, the maxDepth parameter controls the depth of each decision tree, with values like 13 and 16 being explored to strike the right balance between capturing complex patterns and preventing overfitting. We observed during training data with a train-validation split tuning the hyperparameter helps in finding the optimal configuration that maximizes the model's performance.

In terms of evaluation, from our results, we found that the R-squared value for the Train-Validation split (TV) is slightly higher than for Cross-Validation (CV), with a difference of only 0.01. However, when comparing these results on the full dataset, the R-squared values for CV and TV are almost similar. The main difference lies in computation time, with CV taking almost double the time compared to TV on the full dataset.

```
print("R Squared (R2) on test data = %g"
      %rftv_evaluator.evaluate(prediction))
print("RMSE: %f" % rftv_evaluator.evaluate(prediction))

▶ (2) Spark Jobs
R Squared (R2) on test data = 0.681898
RMSE: 86.581514
```
Fig 5 Random Forest regression using Train Validator

```
print("R Squared (R2) on test data = %g"
      %rfcv_evaluator.evaluate(prediction))
print("RMSE: %f" % rfcv_evaluator.evaluate(prediction))

▶ (2) Spark Jobs
R Squared (R2) on test data = 0.671936
RMSE: 89.149534
```
Fig 6 Random Forest Regression using Cross Validator

### 7.2 Gradient Boost Tree Regressor
The results for the Gradient Boost Tree algorithm were applied to both the TrainingValidator function and the CrossValidator function, with results showing similar R2 values of .607 and .587, respectively, applied to the sampled dataset with reasonably well RMSE values of 96.95 and 97.51. The TrainingValidator was selected to be applied on the full dataset as it provided a consistently higher R2 value than the CrossValidator function at nearly half the processing time.

A key finding when GBT was applied with the optimized parameters to the full dataset was a significant increase in performance of 15% increase or (R2 .706) to R2 versus the sample with a similar decrease to RMSE. An additional finding was that the MaxIter parameter had the best performance utilizing the value of 5 consistently, where the default is set to 20.

```
print("R Squared (R2) on test data = %g"
      %gbttv_evaluator.evaluate(prediction))
print("RMSE: %f" % gbttv_evaluator.evaluate(prediction))

▶ (2) Spark Jobs
R Squared (R2) on test data = 0.610171
RMSE: 95.847216
```
Fig 7 Gradient Boost Tree Using Train Validator

```
print("R Squared (R2) on test data = %g"
      %gbtcv_evaluator.evaluate(prediction))
print("RMSE: %f" % gbtcv_evaluator.evaluate(prediction))

▶ (2) Spark Jobs
R Squared (R2) on test data = 0.593359
RMSE: 97.892150
```
Fig 8 Gradient Boost Tree Using Cross Validator

### 7.3 Factorization Machines
Hyperparameter tuning was conducted to determine the optimal configuration of the model, resulting in the selection of the best-performing model. The step size 1 and 0.5 worked well with the model to get higher accuracy.

The R2 and RMSE values were tested on a sample dataset, both for cross-validation and train validation. The results showed that the Train-Validation R2 value is slightly higher, with only a 0.1 difference. On the full dataset, the R-squared values for CV and TV are nearly identical. However, the significant difference between the two lies in the computation time, with cross-validation taking approximately twice as long as train validation.

```
print("R Squared (R2) on test data = %g"
      %fm_evaluator.evaluate(prediction))
print("RMSE: %f" % fmv_evaluator.evaluate(prediction))

▶ (2) Spark Jobs
R Squared (R2) on test data = 0.467902
RMSE: 111.979389
```
Fig 9 Factorization Machine Using Train Validator

```
print("R Squared (R2) on test data = %g"
      %fm_cvevaluator.evaluate(prediction))
print("RMSE: %f" % fm_cvevaluator.evaluate(prediction))

▶ (2) Spark Jobs
R Squared (R2) on test data = 0.458519
RMSE: 111.932804
```
Fig 10 Factorization Machine Using Cross Validator

### 7.4 Decision Tree Regressor
The following results have been obtained from testing the R2 and RMSE values on a sample dataset, both for cross-validation and train validation. On the sample dataset, the cross-validation (CV) R2 value is slightly higher, with a mere 0.02 difference. Hence, we can conclude that the difference is negligible. Allowing the decision tree to grow excessively deep during cross-validation can result in overfitting. To overcome this issue, we can employ a train-validation split instead. This approach enables us to control the tree's depth and improve its generalization capability.

```
print("R Squared (R2) on test data = %g"
%dttv_evaluator.evaluate(prediction))
print("RMSE: %f" % dttv_evaluator.evaluate(prediction))

▶ (2) Spark Jobs
R Squared (R2) on test data = 0.591955
RMSE: 95.414600
```
Fig 11 Decision Tree Regression Train Validator

```
print("R Squared (R2) on test data = %g"
%dtcv_evaluator.evaluate(prediction))
print("RMSE: %f" % dtcv_evaluator.evaluate(prediction))

▶ (2) Spark Jobs

R Squared (R2) on test data = 0.640657
RMSE: 97.050022
```

Fig 12 Decision Tree Regression Using Cross Validator

## 8. FULLDATESET COMPARISON

| Cross Validation | | | |
|---|---|---|---|
| Algorithms | R2 | RMSE | Model Training Time |
| Random Forest | 0.728 | 80.027 | 3 hr 30 mins |
| Gradient Boost Tree | 0.708 | 82.80 | 1 hr 53 mins |
| Decision Tree | 0.63 | 91.99 | 1 hr 30 mins |
| Factorization Machines | 0.46 | 111.02 | 5 hr |
| Train Validation | | | |
| Algorithms | R2 | RMSE | Model Training Time |
| Random Forest | 0.728 | 79.92 | 1 hr 30 mins |
| Gradient Boost Tree | 0.706 | 83.17 | 53 mins |
| Decision Tree | 0.64 | 92.04 | 47 mins |
| Factorization Machines | 0.46 | 111.02 | 3 hr |

Fig 13 Full Dataset Comparison Table

The results and insights obtained by comparing the performance of four algorithms on the Full dataset using both Cross Validation (CV) and Train Validation (TV) are as follows. The comparison results are presented in tables, and the R2 and RMSE values indicate little difference between CV and TV. However, there is a noticeable difference in the training time of the models. The Random Forest and Gradient Boost Tree algorithms are found to have the highest accuracy, both achieving above 70%.

Based on this analysis, the recommended model to use in the Train Validation scenario is the Gradient Boost Tree (GBT), which has a shorter training time compared to other algorithms while maintaining a similar level of accuracy as the Random Forest. This conclusion is supported by the fact that GBT has a shorter training time compared to other algorithms while maintaining a similar level of accuracy as the Random Forest.

## 9. Conclusion

Our paper's objective was to discover the most efficient models for forecasting flight ticket prices. We conducted a comprehensive analysis to predict prices using various Regression algorithms and compared their performance. Out of the four regression algorithms tested, the Gradient Boost Tree algorithm demonstrated the highest accuracy for price prediction, achieving an RMSE of 83.17 and an R2 value of 0.706. Consequently, this algorithm emerged as the most suitable choice for predicting prices compared to the other regression algorithms.

Additionally, our study emphasized the significance of feature importance analysis, which enables informed decisions regarding feature selection, model improvement, and future data collection. By understanding the relative importance of different features, we can enhance the accuracy and efficiency of predictive models. We evaluated the performance of the machine learning models using both cross-validation and train-validation split techniques. Cross-validation was valuable in selecting the optimal hyperparameters, while the train-validation split allowed us to assess the overall model performance on new data.

The implications of our findings have practical applications for both airlines and customers. Airlines can leverage predictive capabilities to optimize pricing strategies for specific routes and seasons. By analyzing pricing trends, airlines can develop effective strategies tailored to different routes and periods. Similarly, customers can utilize the dataset to forecast future flight prices and plan their journeys accordingly. Applying these insights has the potential to enhance the efficiency and effectiveness of pricing strategies in the airline industry, benefiting both businesses and customers alike.

## References

[1] R. R. Subramanian, M. S. Murali, B. Deepak, P. Deepak, H. N. Reddy, and R. R. Sudharsan, "Airline Fare Prediction Using Machine Learning Algorithms,"
IEEE Xplore, Jan. 01, 2022.
https://ieeexplore.ieee.org/abstract/document/9716563

[2] P. S. Singh, P. Samanta, "Flight Fare Prediction System Using Machine Learning," Oct. 20, 2022.
https://dx.doi.org/10.2139/ssrn.4269263

[3] Nitika Verma. (2021). "Regression — flight price prediction." Retrieved from https://medium.com/analytics-vidhya/regression-flight-price-prediction-6771fc4d1fb3

[4] "ML tuning: Model selection and hyperparameter tuning."ML Tuning - Spark 3.4.0 Documentation. (n.d.)
https://spark.apache.org/docs/latest/ml-tuning.html

[5] D. WONG, "Flight Prices", Oct. 2022,
https://www.kaggle.com/datasets/dilwong/flightprices?resource=download